\documentclass[10pt,twocolumn,letterpaper]{article}

\usepackage{iccv}
\usepackage{times}
\usepackage{epsfig}
\usepackage{graphicx}
\usepackage{amsthm}
\usepackage{amsmath}
\usepackage{amssymb}

\usepackage{authblk}

\usepackage[pagebackref=true,breaklinks=true,letterpaper=true,colorlinks,bookmarks=false]{hyperref}
\usepackage{hyperref}
\usepackage{algorithm} 
\usepackage{algorithmic} 
\usepackage{listings}
\usepackage{xcolor}
\usepackage{subfigure}
\usepackage{multirow}
\usepackage{arydshln}
\usepackage{bm}
\usepackage{threeparttable}
\usepackage{booktabs}
\usepackage{verbatim}

\theoremstyle{plain}
\newtheorem{definition}{Definition}
\newtheorem{proposition}{Proposition}

\iccvfinalcopy 

\def\iccvPaperID{4223} 

\ificcvfinal\fi

\begin{document}

\title{Towards Mixed-Precision Quantization of Neural Networks via Constrained Optimization}

\author[1,2]{Weihan Chen}
\author[1]{Peisong Wang}
\author[1\thanks{Corresponding Author}]{Jian Cheng}
\affil[1]{NLPR \& AIRIA, Institute of Automation, Chinese Academy of Sciences}
\affil[2]{School of Artificial Intelligence, University of Chinese Academy of Sciences \authorcr \tt\small chenweihan2018@ia.ac.cn, \{peisong.wang, jcheng\}@nlpr.ia.ac.cn}

\renewcommand\Authsep{ \t }
\renewcommand\Authands{ \t }
\setlength{\affilsep}{0.1em}

\maketitle
\ificcvfinal\thispagestyle{empty}\fi

\begin{abstract}
Quantization is a widely used technique to compress and accelerate deep neural networks.
However, conventional quantization methods use the same bit-width for all (or most of) the layers, which often suffer significant accuracy degradation in the ultra-low precision regime and ignore the fact that emergent hardware accelerators begin to 
support mixed-precision computation. 
Consequently, we present a novel and principled framework to solve the mixed-precision quantization problem in this paper. 
Briefly speaking, we first formulate the mixed-precision quantization as 
a discrete constrained optimization problem. 
Then, to make the optimization tractable, we approximate the objective function with second-order Taylor expansion and propose an efficient approach to compute its Hessian matrix. Finally, based on the above simplification, we show that the original problem can be reformulated as a Multiple-Choice Knapsack Problem (MCKP) and propose a greedy search algorithm to solve it efficiently. 
Compared with existing mixed-precision quantization works, our method is derived in a principled way and much more computationally efficient. 
Moreover, extensive experiments conducted on the ImageNet dataset and various kinds of network architectures also demonstrate its superiority over
existing uniform and mixed-precision quantization approaches.
\end{abstract}

\section{Introduction}
\label{sec:intro}

In the past few years, Convolutional Neural Networks (CNNs) have been leading
new state-of-the-art in almost every computer vision tasks, ranging from image 
classification \cite{alexnet, vgg, resnet}, segmentation \cite{fcn, deeplabv1, segnet}, and object detection \cite{faster-rcnn, fpn, ssd}. However, such performance boosts often come at 
the cost of increased computational complexity and storage overhead. 
In many real-time applications, 
storage consumption and latency are crucial, which on the other hand, have posed great challenges to the deployment of these networks. Under this circumstance, a variety of methods have been proposed, including low-rank decomposition \cite{accel_deep, linear_struct}, knowledge distillation \cite{kd, fitnet}, low-precision quantization \cite{google-quantize, HWGQ}, filter pruning \cite{filter-pruning, rethinking}, etc, to achieve CNNs compression and acceleration.

Among these approaches, quantization becomes one 
of the most hardware-friendly one by approximating real-valued weights and activations with lower bit-width fixed-point representations. 
Meanwhile, network inference can be performed using cheaper fixed-point multiple-accumulation (MAC) operations. 
As a result, we can significantly reduce the storage overhead and inference latency of CNNs. 

Most of the existing quantization methods \cite{HWGQ, dorefa-net, lq-net, pact, abc-net, twn, xnor-net, ttq, inq} use the same bit-width for all (or most of) the layers. Such a uniform bit-width assignment can be suboptimal from two aspects. First, different layers have different redundancy and contribute differently to the final performance. Therefore, uniformly quantizing a network to ultra-low precision often leads to significant accuracy degradation.
Second, emergent hardware accelerators, such as BISMO \cite{bismo} and BitFusion \cite{bit-fusion}, begin to support mixed-precision computation for greater flexibility.
Consequently, to achieve a better trade-off between accuracy and efficiency, there is a rising demand to apply mixed-precision quantization by finding the optimal bit-width for each layer. 

However, mixed-precision quantization is difficult for two reasons. 
First, the search space of choosing bit-width assignment is huge. 
For a network with $N$ layers and $M$ candidate bit-widths in each layer, an exhaustive combinatorial search has exponential time complexity ($\mathcal{O}(M^N)$). 
Second, to evaluate the performance of each bit-width assignment truly, we need to finetune the quantized network until it converges, which may take days for the large-scale dataset. 
Therefore, a large bulk of mixed-precision quantization methods \cite{haq, hawq, hawq-v2, dnas, autoq, good-para} have been proposed recently to solve the problem approximately.
Based on different approximation strategies, we can categorize these methods roughly into two groups as discussed below.

\textbf{Search-Based: }
To reduce the computation complexity, search-based methods aim to sample more efficiently and obtain enough performance improvement with only a small number of evaluations. 
Therefore, HAQ \cite{haq} leverages reinforcement learning to determine the quantization policy layer-wise and take the hardware accelerator's feedback in the design. After that, AutoQ \cite{autoq} proposes a hierarchical-DRL-based technique to search for the bit-width kernel-wise. Furthermore, EvoQ \cite{evoq} alters to employ the evolutionary algorithm with limited data. 
Generally speaking, as the time cost of performance evaluation is still huge, search-based methods limit the exploration of search space greatly to make the algorithms computationally feasible. 

\textbf{Criterion-Based: }
Differently, criterion-based methods instead aim to reduce the time cost of  performance evaluation through kinds of criteria that are easy to compute.
Among them, HAWQ \cite{hawq} utilizes the top Hessian eigenvalue as the measure of quantization sensitivity of each layer. 
Although provided with relative sensitivity, it still requires a manual selection of the bit-width assignment. To solve the problem, HAWQ-V2 \cite{hawq-v2} proposes a Pareto frontier based method to finish it automatically and alters to take the trace of Hessian matrix as the criterion.  
Although effective in practice, most existing criteria are still ad-hoc and lack of principled explanation for the optimality. 

Overall, mixed-precision quantization remains an open problem so far given its intrinsic difficulty. In this paper, we present a novel and principled framework to solve it. Specifically, we first formulate mixed-precision quantization as a discrete constrained optimization problem with regard to the bit-width assignment among layers, which provides a principled and holistic view for our further analysis. As it is intractable to calculate the original objective function, we then approximate it with Taylor expansion and propose an efficient approach to compute the Hessian matrix of each layer. 
Finally, based on the above simplification, we show that the original problem can be reformulated as a special variant of the Knapsack problem called Multiple-Choice Knapsack Problem (MCKP) and 
propose a greedy search algorithm to solve it efficiently. 

Compared with existing works, our method is first computationally efficient and even significantly faster than criterion-based approaches. 
Take ResNet50 as an example, it only takes less than 2 minutes to finish the whole bit-width assignment procedure with a single RTX 2080Ti.
Please refer to the \textbf{Efficiency Analysis} in Section \ref{sec:method analysis} for details. 
Second, as our method is derived in a principled way, it is more interpretable compared with other ad-hoc ones and also accessible for further improvement such as a more sophisticated solving algorithm. 
Third, our method achieves a better trade-off between search-based and 
criterion-based methods. 
Compared with search-based ones (e.g. HAQ), our method reduces the evaluation cost greatly to search for the optimal bit-width assignment in a much larger space. 
Compared with criterion-based ones (e.g. HAWQ), our method is based on the whole Hessian matrix instead of the eigenvalues only. 
Empirically, extensive experiments conducted on the ImageNet dataset and various kinds of networks justify the superiorities of our method over them. 

To summarize, our main contributions are three-fold:
\begin{itemize}
    \item We first formulate the mixed-precision quantization as a discrete 
    constrained optimization problem to provide a principled and holistic 
    view for further analysis.

    \item To solve the optimization, we propose an efficient 
    approach to compute the Hessian matrix and then reformulate it 
    as Multiple-Choice Knapsack Problem (MCKP) to be solved by greedy search efficiently
    
    \item Extensive experiments are conducted to demonstrate the efficiency and
	effectiveness of our method over other uniform/mixed-precision 
	quantization ones.
\end{itemize}


\section{Related Work}
\label{sec:related-work}
As convolutional neural networks often suffer from significant
redundancy in their parameterization, lots of works have emerged and focus on the acceleration and compression of CNNs recently. 
Here we only review the works related to ours and refer the reader to recent 
surveys \cite{quantization-survey, chengyu, vivienne, chengjian, hansong} for a comprehensive overview. 

Full-precision parameters are not required in achieving high performance in CNNs. To compress the models, \cite{inq} proposed to quantize the weights incrementally and showed that with reduced precision to 2-5 bits classification accuracy on the ImgeNet could be even slightly higher. 
Furthermore, several recent works \cite{pact, dorefa-net, HWGQ} focused on quantizing both the weights and activations for acceleration gain. 
As conventional quantization methods use the same bit-width for all (or most of) the layers and often 
suffer significant accuracy degradation in the ultra-low precision regime, lots of different methods have been proposed to address it through mixed-precision quantization recently. 
As we have discussed in Section \ref{sec:intro}, most existing works can be empirically categorized into two groups, namely search-based and criterion-based ones. 
Besides, \cite{dnas} formulate the problem as a neural architecture search problem and propose a differential neural architecture search (DNAS) framework to efficiently explore the search space with gradient-based optimization. 
\cite{good-para} proposes to parametrize the quantizer with step size and dynamic range which are optimized through straight-through estimator (STE) \cite{ste}, and then the bit-width of each layer can be inferred from them automatically.

\section{Methodology}
\label{sec:methodology}
In this section, we firstly introduce a general formulation of 
mixed-precision quantization as a discrete constrained optimization problem with regard to the bit-width assignment. Secondly, as it is intractable to calculate the original objective function, we approximate it with second-order Taylor expansion and propose an efficient approach to compute its Hessian matrix. Finally, we transform the optimization into a special variant of the Knapsack problem called Multiple-Choice Knapsack Problem (MCKP) and propose a greedy search algorithm to solve it efficiently.

\subsection{Notation and Background}
\textbf{Notation: }We assume a $L$-layer Convolutional Neural Network 
$f: \Omega \times \mathbb{X} \rightarrow \mathbb{Y}$ and a training dataset 
of $N$ samples $(\mathbf{x}^{(n)}, \mathbf{y}^{(n)}) \in \mathbb{X} \times \mathbb{Y}$ 
with $n = 1, \dots, N$. The model maps each sample $\mathbf{x}^{(n)}$ to a 
prediction $\hat{\mathbf{y}}^{(n)}$ using some paramaters $\theta \in \Omega$. 
Then the predictions are compared with the ground truth $\mathbf{y}^{(n)}$ and 
evaluated with a task-specific loss function $\ell: \mathbb{Y} \times \mathbb{Y} \rightarrow \mathbb{R}$, for example the cross-entropy loss for image classification. 
This leads to the objective function to minimize $\mathcal{L}: \Omega \rightarrow \mathbb{R}$, 
\begin{equation}
\begin{aligned}
\mathcal{L}(\theta) &= \frac{1}{N} \sum_{n=1}^N \ell(f(\theta, \mathbf{x}^{(n)}), \mathbf{y}^{(n)}) \\
&= \frac{1}{N} \sum_{n=1}^N \ell^{(n)}(\theta). \\
\end{aligned}
\end{equation}
Specially, for the $l$-th convolutional or full-connected layer, we denote its 
weight tensor as $W^{(l)} \in \mathbb{R}^{c_o \times c_{i} \times k \times k}$ 
and its flattened version as $w^{(l)} \in \mathbb{R}^{c_o c_{i} k^2}$, where $k$ is 
the kernel size (equals to $1$ for full-connected layers), $c_i$ and $c_o$ are the number of input and output channels, respectively.

\textbf{Quantization Background: }
The purpose of quantization is to map the floating-point values into a finite set with discrete elements. 
Mathematically, we can formulate the quantization 
function as $Q: \mathbb{R}^{D} \times \mathbb{Z}^{+} \rightarrow \Pi_{b}$, which 
takes full-precision vector and quantization bit-width as input and outputs 
the quantized vector. 
In this paper, we only consider 
uniform symmetric quantization as it takes little extra overhead to implement 
in most hardware platforms. 
As a result, $\Pi_{b}$ equals to $s \times \{ -2^{b-1}, \dots, 0, \dots, 2^{b-1}-1 \}$ 
for signed input and $s \times \{ 0, \dots, 2^{b}-1 \}$ for unsigned 
one, where $b$ is the quantization bit-width and $s$ is the step size between two 
consecutive grid points. Here we adopt Minimum Squared Error (MSE) as the quantization criterion and solve the following minimization problem
\begin{equation}
\mathop{\min}_{s} \ \lVert w - Q(w, b) \rVert_{2} \quad s.t. \ Q(w, b) \in \Pi_{b}
\end{equation}
to get the step size $s$ with a given bit-width $b$. After that, one can easily get the quantized vector by leveraging the rounding-to-nearest operation, e.g. 
$Q(w, b) = clip(\lfloor w/s \rceil, 0, 2^{b-1}) \times s$ for unsigned input.

\subsection{Problem Formulation}
Let $w := \{ w^{(l)} \}_{l=1}^{L}$ be the set of flattened weight tensors of a CNN which has $L$ layers. To find the optimal bit-width assignment with the goal of compression or acceleration, we have the following discrete constrained problem:
\begin{equation}
\label{eq:formulation}
\begin{aligned}
\mathop{\min}_{\{ b^{(l)} \}_{l=1}^{L}} & \frac{1}{N} \sum_{n=1}^{N} \ell(f(w+\Delta w, \mathbf{x^{(n)}}), \mathbf{y^{(n)}}) \\
\mbox{s.t.} \quad & \Delta w^{(l)} = Q(w^{(l)}, b^{(l)}) - w^{(l)} \\
& \quad \mathcal{C}_{j}(b^{(1)}, \dots, b^{(L)}) \le 0           \\
& \quad \quad  \quad b^{(l)} \in \mathbb{B}                        \\
l \in & \{1, \dots, L\}, j \in \{1, \dots, M\}     \\
\end{aligned}
\end{equation}
Problem (\ref{eq:formulation}) is a general form of mixed-precision quantization. More specifically, inequality constraints $\mathcal{C}_j$ for 
$j \in \{1, \dots, M\}$ indicate our quantization budgets, such as model compression, flops reduction or both of them. 
For fair comparison with other mixed-precision methods, here we consider the constraint of model compression. That is to say, we instantiate the 
quantization target as 
\begin{equation}
\label{eq:constraint}
\sum_{l=1}^{L} \lvert w^{(l)} \rvert \cdot b^{(l)} \le b_{target} \cdot \sum_{l=1}^{L} \lvert w^{(l)} \rvert
\end{equation}
where $b_{target}$ denotes our target average bit-width of the network, and 
$\lvert \cdot \rvert$ denotes the length of corresponding vector. 
However, objective function (\ref{eq:formulation}) is computationally expensive as we need 
to evaluate the network on the whole training dataset for each candidate 
bit-width assignment. Instead, it is replaced with the 
second-order Taylor expansion
\begin{equation}
\begin{aligned}
\mathcal{L}(w + \Delta w) &= \frac{1}{N} \sum_{n=1}^{N} \ell^{(n)}(w + \Delta w) \\
&\approx \mathcal{L}(w) + g_{w}^{T}\Delta w + \frac{1}{2}\Delta w^{T}H_{w} \Delta w.
\end{aligned}
\end{equation}
Here we use $g_{w} := \nabla\mathcal{L}(w)$ and $H_{w} := \nabla^2 \mathcal{L}(w)$ to denote the first-order gradient and 
second-order Hessian matrix respectively. 
First, the zero-order term is a constant which can be removed without any 
influence on the optimization. 
Then, given a pre-trained model, it's reasonable to assume that it has converged to a local miniumum with nearly zero gradient vector. 
Therefore, we conclude with the only reserved term $\Delta \mathcal{L} = \frac{1}{2}\Delta w^{T}H_{w} \Delta w$, which is our final objective function that approximates the loss perturbation from quantization. 
However, although the gradient can be computed in linear time, the Hessian matrix is much harder to compute and store as its complexity is quadratic to the number of parameters. 
Hence, we need to find an efficient approach to compute and store these matrices.

\subsection{Approximated Hessian Matrix}
Denote the neural network output of each sample as $f^{(n)}(w) = [f^{(n)}_1(w), \cdots, f^{(n)}_p(w)]^T \in \mathbb{R}^p$. 
According to the chain rule, the Hessian matrix can be computed by
\begin{equation}
\label{eq:hessian}
\begin{aligned}
\frac{\partial^{2} \mathcal{L}}{\partial w_i w_j} &= \frac{1}{N} \sum_{n=1}^{N}
\frac{\partial^2 \ell^{(n)}}{\partial w_i w_j} \\
&= \frac{1}{N} \sum_{n=1}^{N} \biggl( \frac{\partial}{\partial w_j} \Bigl( \sum_{k=1}^{p} \frac{\partial \ell^{(n)}}{\partial f^{(n)}_k}  \frac{\partial f^{(n)}_k}{\partial w_i} \Bigr) \biggr) \\
&= \frac{1}{N} \sum_{n=1}^{N} \sum_{k=1}^{p} \frac{\partial \ell^{(n)}}{\partial f^{(n)}_k} \frac{\partial^2 f^{(n)}_k}{\partial w_i w_j} \\ 
&+ \frac{1}{N} \sum_{n=1}^{N} \sum_{k,l=1}^{p} \frac{\partial f^{(n)}_k}{\partial w_i} \frac{\partial^2 \ell^{(n)}}{\partial f^{(n)}_k \partial f^{(n)}_l} \frac{\partial f^{(n)}_l}{\partial w_j}.
\end{aligned}
\end{equation}
We note that the first term of Eq. (\ref{eq:hessian}) is the bottleneck of computation cost.
To calculate the Hessian efficiently, we approximate it by neglecting this term  (\emph{see supplementary material for the theoretical\&empirical analysis of this approximation}). 
To simplify the notations, we introduce $\nabla f^{(n)}(w) \in \mathbb{R}^{p \times d}$ which is the Jacobian matrix of $f^{(n)}$ on $w$, 
and $\Sigma^{(n)} \in \mathbb{R}^{p \times p}$ which is the Hessian matrix of $\ell^{(n)}$ on $f^{(n)}$. Therefore, the approximated Hessian matrix can be written in matrix form as
\begin{equation}
\tilde{H}_w = \frac{1}{N} \sum_{n=1}^{N}  \nabla^T f^{(n)}(w) 
\Sigma^{(n)} \nabla f^{(n)}(w)
\end{equation}
Then we substitute the Hessian matrix with our approximation into the loss perturbation $\Delta \mathcal{L}$ and get
\begin{equation}
\label{eq:approx}
\begin{aligned}
\Delta \mathcal{L} &= \frac{1}{2}\Delta w^{T}H_{w} \Delta w \approx  \frac{1}{2}\Delta w^{T} \tilde{H}_{w} \Delta w \\
&= \frac{1}{2}\Delta w^{T} \cdot \frac{1}{N} \sum_{n=1}^{N} \nabla^T f^{(n)} 
\Sigma^{(n)} \nabla f^{(n)} \cdot \Delta w \\
&= \frac{1}{2N} \sum_{n=1}^{N} [ \nabla f^{(n)} \Delta w ]^T \Sigma^{(n)} 
[\nabla f^{(n)} \Delta w].
\end{aligned}
\end{equation}
As we can see from Eq. (\ref{eq:approx}), it only involes first-order 
derivate except $\Sigma^{(n)}$ which can be solved analytically with the given loss function. Here we consider the commonly-used loss function in classification task, cross-entropy loss,
\begin{equation}
\mathcal{L}(w) = - \frac{1}{N} \sum_{n=1}^{N} \sum_{k=1}^{p} y^{(n)}_{k} \ \text{log}
f^{(n)}_k
\end{equation}
and it's easy to derive that
\begin{equation}
\Sigma^{(n)} = diag(y^{(n)}_1 / [f^{(n)}_1]^2, \dots, y^{(n)}_p / [f^{(n)}_p]^2).
\end{equation}
Then, it is noted that the ground-truth label $\mathbf{y}^{(n)}$ of each sample 
is a one-hot vector. As a result, we can rewrite the formula (\ref{eq:approx}) as
\begin{equation}
\label{eq:reduced}
\Delta \mathcal{L} = \frac{1}{2N} \sum_{n=1}^{N} \frac{1}{[f^{(n)}_{t^{\ast}}]^2} (\nabla f^{(n)}_{t^{\ast}} \Delta w)^2,
\end{equation}
where $t^{\ast}$ and $\nabla f^{(n)}_{t^{\ast}}$ denote the ground-truth label and $t^{\ast}$-th row of $\nabla f^{(n)}$ respectively. 
It means that we only need to calculate one single row of the Jacobian matrix $\nabla f^{(n)}(w)$ to figure out the loss perturbation $\Delta \mathcal{L}$ of each sample. 
What's more, refer to the \textbf{Convergence Analysis} in Section \ref{sec:method analysis}, the result of loss perturbation converges rapidly as the number of images increases. 
Hence there is no need to traverse the entire dataset, which improves the computation efficiency further. 

\subsection{MCKP Reformulation}
Up to now, we are able to calculate the loss perturbation incurred from the 
quantization of specific bit-with assignment efficiently. To finish the bit-width assignment automatically, we make the assumption that the Hessian matrix is block-diagonal with non-zero terms only within each layer parameters, namely the quantization of different layers is independent of each other. Hence we can reformulate the 
objective function as
\begin{equation}
\label{eq:objective}
\begin{aligned}
\Delta \mathcal{L} &= \frac{1}{2} \Delta w^{T} \tilde{H}_{w} \Delta w \\
& \approx \frac{1}{2} \sum_{l=1}^{L} (\Delta w^{(l)})^{T} \tilde{H}_{w^{(l)}} \Delta w^{(l)}.
\end{aligned}
\end{equation}
Now combine Eq. (\ref{eq:formulation}), (\ref{eq:constraint}) and (\ref{eq:objective}), finally we can reformulate the optimization problem as
\begin{equation}
\label{eq:formulation_final}
\begin{aligned}
\mathop{\min}_{\{ b^{(l)} \}_{l=1}^{L}} & \ \ \frac{1}{2} \sum_{l=1}^{L} (\Delta w^{(l)})^{T} \tilde{H}_{w^{(l)}} \Delta w^{(l)} \\
\mbox{s.t.} \quad & \Delta w^{(l)} = Q(w^{(l)}, b^{(l)}) - w^{(l)} \\
\sum_{l=1}^{L} & \lvert w^{(l)} \rvert  \cdot b^{(l)}  \le b_{target} \cdot \sum_{l=1}^{L} \lvert w^{(l)} \rvert \\
& \quad \quad  \quad b^{(l)} \in \mathbb{B}                        \\
& \quad \quad l \in \{1, \dots, L\}     \\
\end{aligned}
\end{equation}
To solve the problem, we will introduce a special variant of the Knapsack problem 
called Multiple-Choice Knapsack Problem (MCKP) \cite{knapsack-problem} and show that problem (\ref{eq:formulation_final}) can be written as an MCKP.
\begin{definition}
\label{def:MCKP}
Given $k$ classes $N_1, \dots, N_k$ of items to pack in some knapsack of 
capacity $c$. Each item $j \in N_i$ has a profit $\rho_{ij}$ and a weight 
$\omega_{ij}$, and the problem is to choose one item from each class such that the profit sum is maximized without having the weight sum to exceed $c$. The 
Multiple-Choice Knapsack Problem (MCKP) may thus be reformulated as:
\begin{equation}
\begin{aligned}
\mathop{\max}_{x_{ij}} \quad z &= \sum_{i=1}^k \sum_{j \in N_i} \rho_{ij}x_{ij} \\
\mbox{s.t.} \quad & \sum_{i=1}^k \sum_{j \in N_i} \omega_{ij}x_{ij} \le c \\
&  \sum_{j \in N_i} x_{ij} = 1, \ \ x_{ij} \in \{0, 1\} \\
i & \in \{1, \dots, k\}, \ \ j \in N_i.
\end{aligned}
\end{equation}
All coefficients $\rho_{ij}$, $\omega_{ij}$ and $c$ are positive real numbers, and the 
classes $N_1, \dots, N_k$ are mutually disjoint, class $N_i$ having size 
$n_i$. The total number of items is $n = \sum_{i=1}^{k} n_i$. 
\end{definition}
It's evident that problem (\ref{eq:formulation_final}) can be reformulated as an 
instance of MCKP according Definition \ref{def:MCKP}. More specially, each 
class is defined by each layer with size $n_i = \lvert \mathbb{B} \rvert$ which 
denotes the number of candidate bit-width. Then the bit-width assignment of each 
layer can be regarded as an MCKP item. Besides, we define $\omega_{ij}$ as 
$\lvert w^{(i)} \rvert \cdot j$ and $\rho_{ij}$ as 
\begin{equation}
-\frac{1}{2}(\Delta w^{(i)}_j)^{T} \tilde{H}_{w^{(i)}} \Delta w^{(i)}_j
\end{equation}
with $\Delta w^{(i)}_j = Q(w^{(i)}, j) - w^{(i)}$. 
The capacity of knapsack $c$ is our target model size, namely $b_{target} \cdot \sum_{l=1}^{L} \lvert w^{(l)}$, and $x_{ij}$ indicates whether choose bit-width $j$ for layer $i$.

\begin{algorithm} 
\caption{Constrained Optimization-based Algorithm for Mixed-Precision Quantization} 
\label{alg:summary} 
\begin{algorithmic}[1] 
\REQUIRE training dataset $\{ (\mathbf{x}^{(n)}, \mathbf{y}^{(n)} )\}_{n=1}^{N}$,

	     \ \ \quad pre-trained network with weights $\{ W ^{(l)}\}_{l=1}^{L}$,
	     
		 \ \ \quad candidate bit-widths of each layer $\mathbb{B}$,
		      
	     \ \ \quad target average bit-width $b_{target}$
 
\ENSURE bit-width assignment of each layer $\{ b^{(l)} \}_{l=1}^{L}$

\STATE  /* Step 1: \textit{calculate $\Delta w$ of the given network} */
\STATE calculate $\{ \{ \Delta w^{(l)}_{b} = Q(w^{l}, b) - w^{(l)} \}_{b\in\mathbb{B}} \}_{l=1}^L$

\STATE  /* Step 2: \textit{calculate $\Delta \mathcal{L}$ of the given network} */
\STATE initialize loss perturbation $\{ \{ \Delta \mathcal{L}_{b}^{(l)} \}_{b\in\mathbb{B}} \}_{l=1}^{L}$ with zero
\FOR{$n=1$ \TO $N$} 
\STATE compute output and gradient for $(\mathbf{x}^{(n)}, \mathbf{y}^{(n)} )$
\STATE update $\{ \{ \Delta \mathcal{L}_{b}^{(l)} \}_{b\in\mathbb{B}} \}_{l=1}^{L}$ according to Eq. \ref{eq:reduced}
\ENDFOR


\STATE  /* Step 3: \textit{greedy search to solve MCKP problem} */

\STATE  /* Step 3.1: \textit{eliminate dominated items of each class} */
\FOR{$l=1$ \TO $L$}
\STATE remove the dominated items based on $\{ \Delta \mathcal{L}^{(l)}_b\}_{b\in\mathbb{B}}$ and update the candidate bit-widths denoted by $\mathbb{B}^{(l)}$
\ENDFOR

\STATE  /* Step 3.2: \textit{assign bit-width with greedy criterion} */
\STATE initialize $b^{(l)}$ with the minimum bit-width of $\mathbb{B}^{(l)}$

\WHILE{average bit-width below the target $b_{target}$}
\FOR{$l=1$ \TO $L$} 
\STATE obtain the next available bit-width $\hat{b}^{(l)}$ and its corresponding 
loss perturbation $\Delta \mathcal{L}_{\hat{b}^{(l)}}^{(l)}$
\STATE calculate the priority of layer as $\frac{\Delta \mathcal{L}_{\hat{b}^{(l)}}^{(l)} - \Delta \mathcal{L}_{b^{(l)}}^{(l)}}{(\hat{b}^{(l)} - b^{(l)})\cdot \lvert w^{(l)} \rvert}$
\ENDFOR
\STATE sort the priority among layers, denote the largest one as layer $l^{\star}$ and its bit-width as $\hat{b}^{(l^{\star})}$
\STATE update the bit-width assignment by $b^{(l^{\star})} \leftarrow \hat{b}^{(l^{\star})}$
\ENDWHILE
\end{algorithmic} 
\end{algorithm}

As MCKP is NP-hard, 
here we propose a greedy search algorithm to solve it efficiently. To this end, we first introduce some fundamental properties of MCKP.
\begin{definition}
\label{def:dominance}
If two items $r$ and $s$ in the same class $N_i$ satisfy that
\begin{equation}
\omega_{ir} \le \omega_{is} \quad \text{and} \quad \rho_{ir} \ge \rho_{is},
\end{equation}
then we say that item $r$ dominates item $s$.
\end{definition}
\noindent
Then it is easy to get the following conclusion.
\begin{proposition}
\label{prop:dominace}
Given two items $r,s \in N_i$. If item $r$ dominates item $s$ then an optimal 
solution to MCKP with $x_{is} = 0$ exists.
\end{proposition}
As a consequence, we only have to consider the undominated items in the solution of MCKP. 
Briefly speaking, we first filter all the dominated items and then initialize each layer with the minimum available bit-width. After that, each time we choose the layer with the highest priority based on our proposed greedy criterion and increase its bit-width until the target compression constraint is broken. 
In the end, the overall procedure of our proposed method is summarized in Algorithm \ref{alg:summary}, please refer to it for details of implementation.


\begin{figure*}[htbp]
\centering
\subfigure[ResNet-18]{\label{fig:convergence:resnet18}\includegraphics[scale = 0.24]{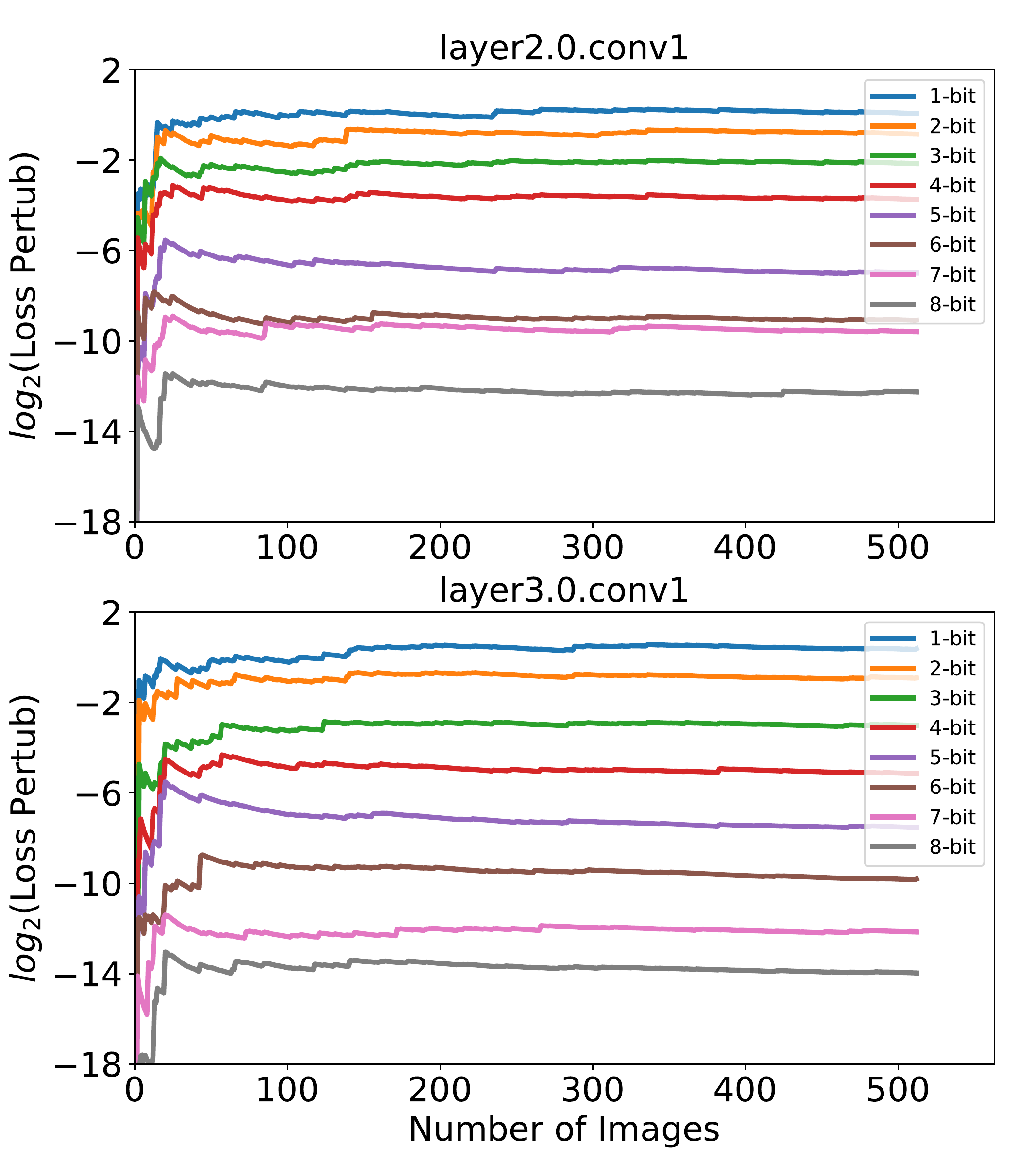}}
\subfigure[ResNet-50]{\label{fig:convergence:resnet50}\includegraphics[scale = 0.24]{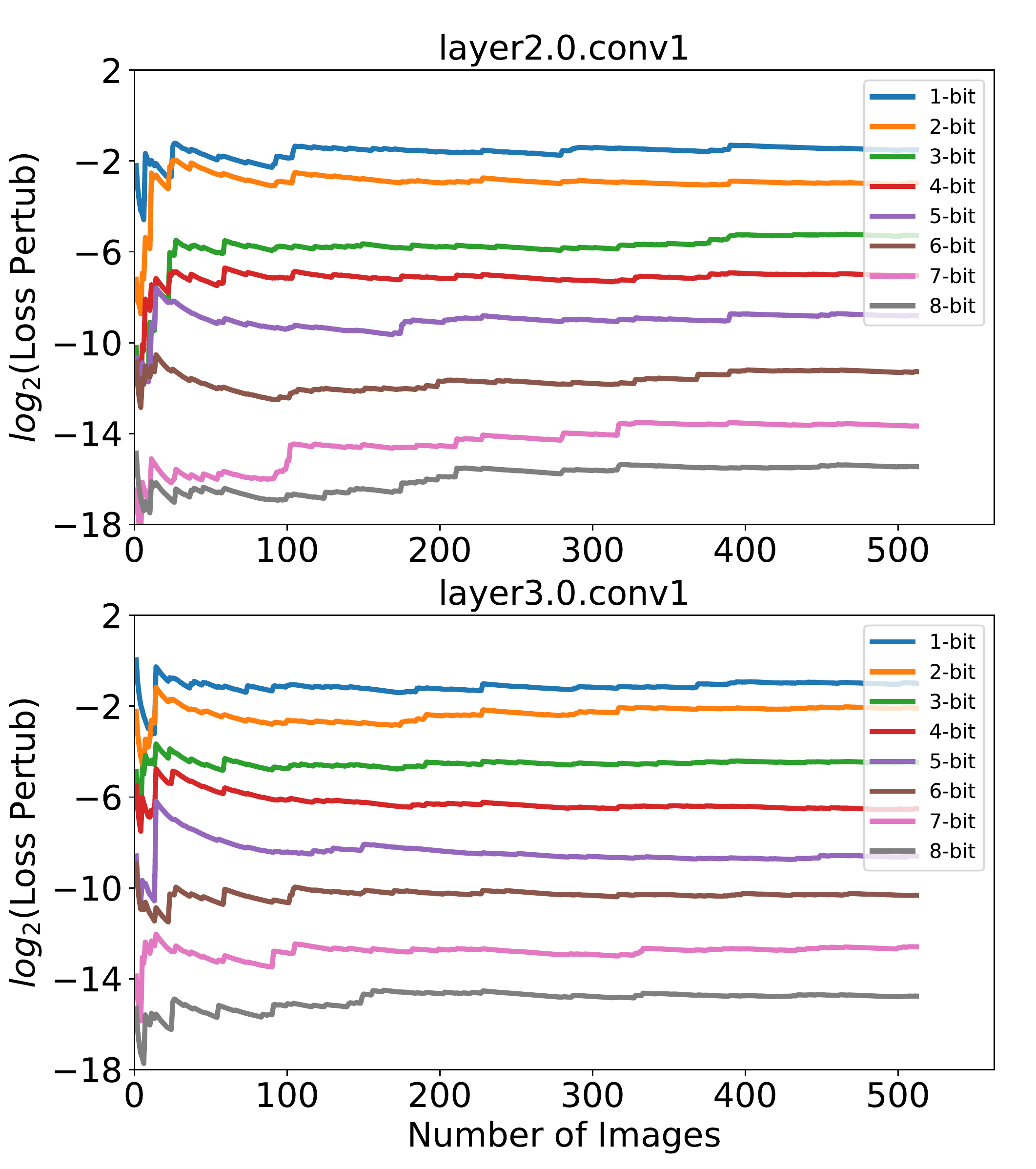}}
\subfigure[Mobilenet-V2]{\label{fig:convergence:mobilenetv2}\includegraphics[scale = 0.24]{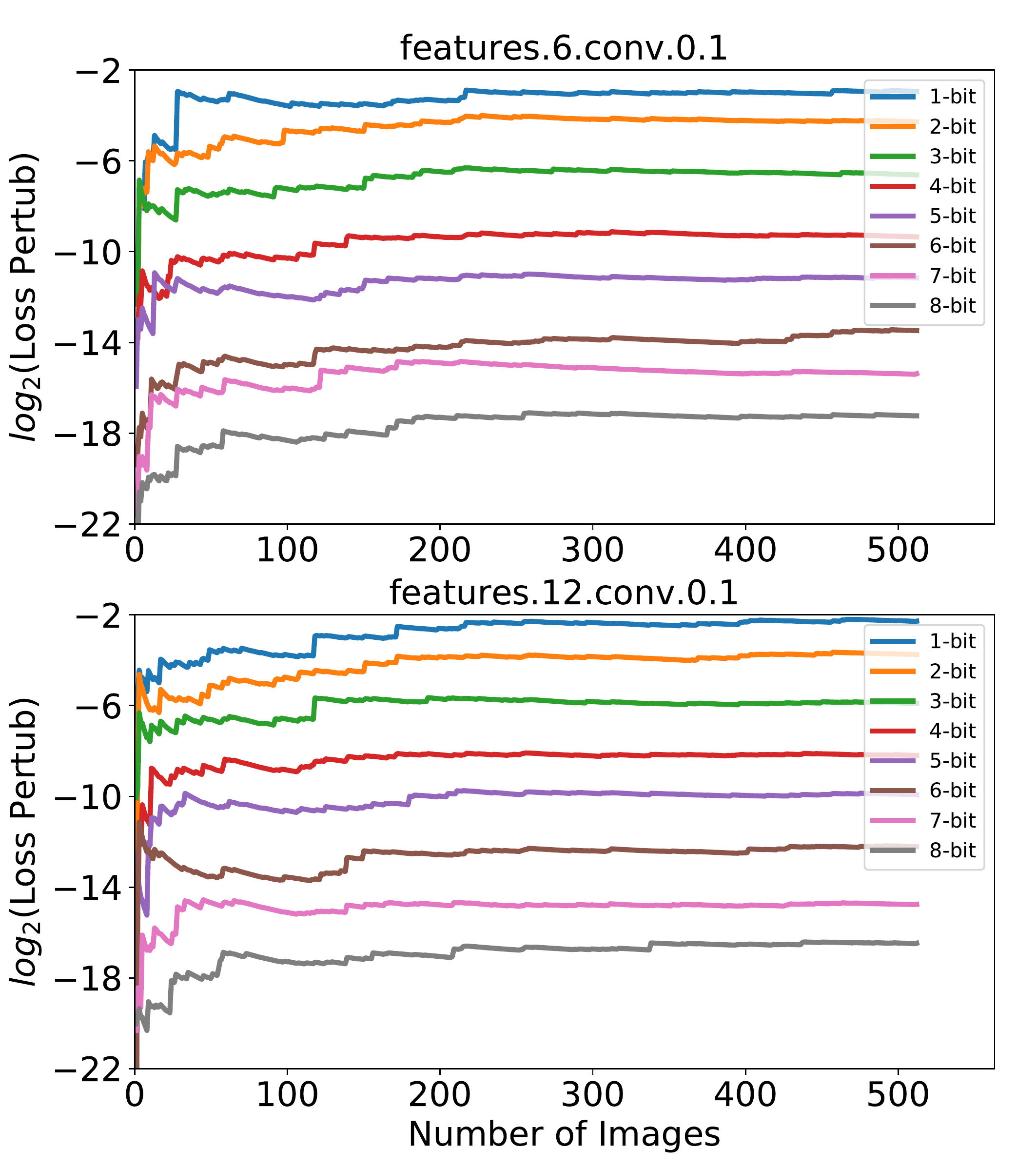}}
\caption{Relationship between the convergence of loss perturbation and the number of images for various kinds of architectures.}
\label{fig:convergence}
\end{figure*}

\section{Experiments}
\label{sec:experiments}

\begin{figure}[htbp]
\centering
\includegraphics[scale = 0.15]{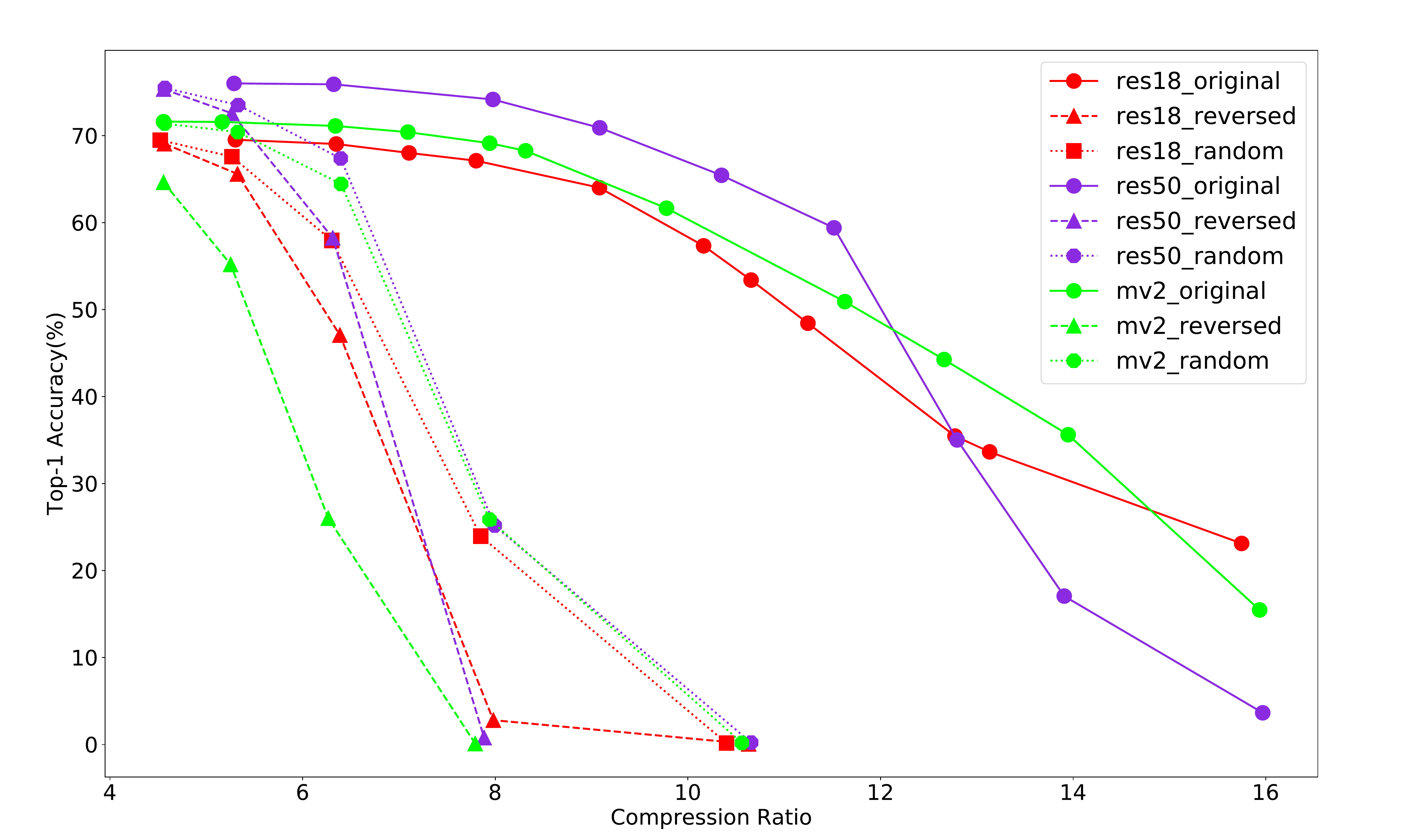}
\caption{Comparision of different criteria for greedy search of various kinds of architectures.}
\label{fig:greedy_ablation}
\end{figure}

\subsection{Method Analysis}
\label{sec:method analysis}
In this section, we conduct comparative analysis from different aspects to understand our method further. 

\textbf{Convergence Analysis: }As shown in Algorithm \ref{alg:summary}, we need to traverse the entire given training dataset for the calculation of loss perturbation $\Delta \mathcal{L}$. However, as the scale of the dataset increases, this calculation will become the bottleneck of time cost of the whole algorithm. 
Therefore, we first analyze the convergence of loss perturbation with regard to the number of images. 
As shown in Figure \ref{fig:convergence}, the results converge rapidly with a few hundred images for all kinds of architectures, which indicates the chance to improve the algorithm's efficiency significantly. 
Consequently, we only sample 1024 images randomly to figure out the loss perturbation in the following experiments. 

\textbf{Efficiency Analysis: }Although most previous methods focus on performance improvement, we argue that computation efficiency should also be taken seriously in the actual deployment. 
For the search-based methods of AutoQ \cite{autoq}, it explores 400 episodes totally with the proposed hierarchical-DRL algorithm and fine-tunes each quantization policy with ten epochs in the randomly selected 100 categories of images from ImageNet for evaluation. In other words, it takes more than 1000 GPU-hours of RTX 2080Ti to search for the optimal bit-width of ResNet-50. 
And for the criterion-based methods of HAWQ-V2 \cite{hawq-v2}, which is much more efficient, it still needs 30 minutes with 4 GPUs to calculate all the average Hessian traces of ResNet-50, let alone the time cost of the Pareto frontier calculation for automatic bit-width assignment. 
By contrast, thanks to the rapid convergence of loss perturbation and efficient greedy search algorithm, our method \emph{only takes less than 2 minutes to finish the whole bit-width assignment procedure of ResNet-50 with a single RTX 2080Ti} and demonstrates significant efficiency advantage (\emph{see supplementary material for the theoretical analysis of computation complexity}). 

\subsection{Ablation Study}
In this section, we conduct ablation studies to justify the effectiveness of our method's different parts.

\begin{figure*}[htbp]
\centering
\subfigure[ResNet-18]{\label{fig:compression:resnet18}\includegraphics[scale = 0.185]{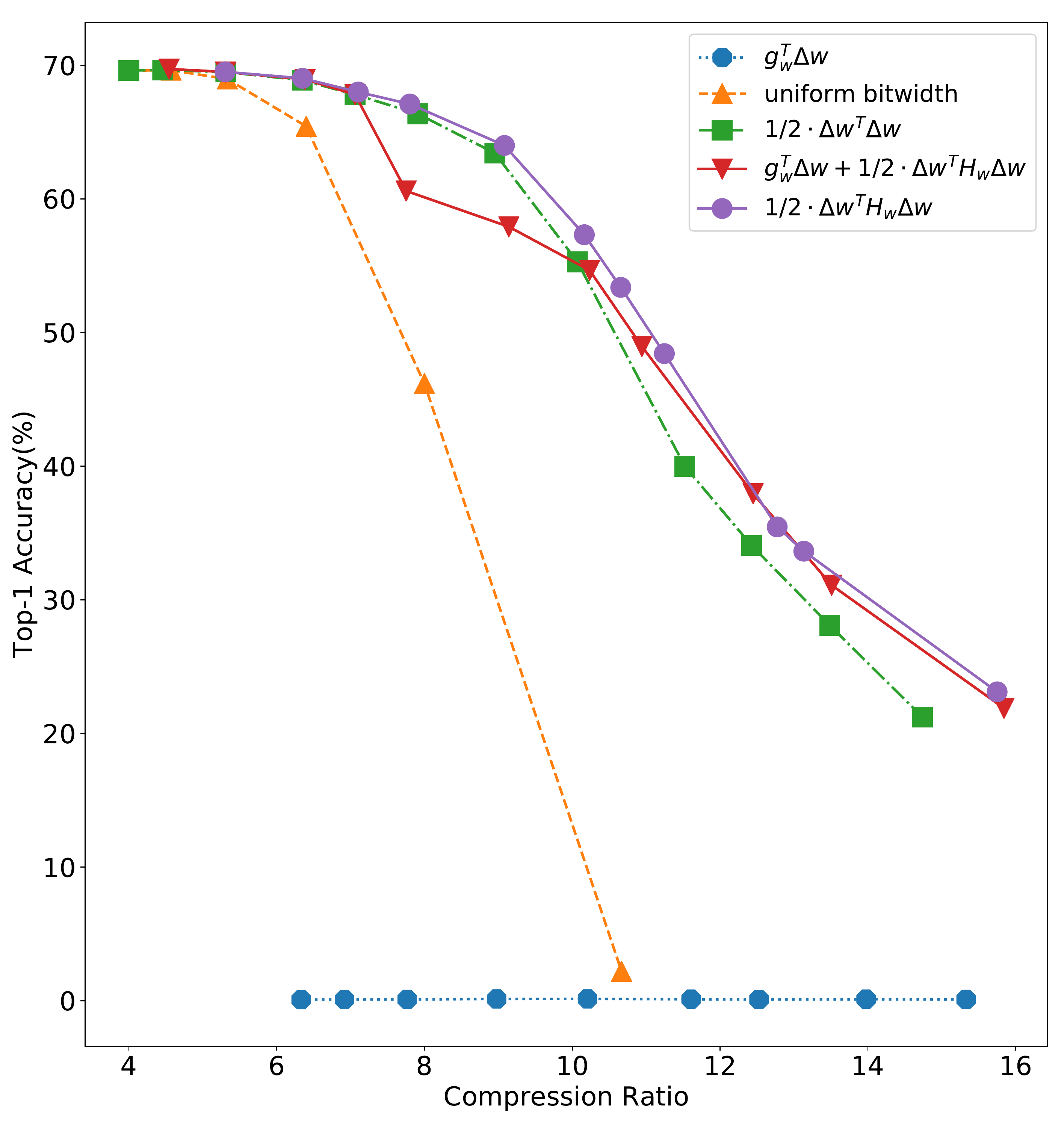}}
\subfigure[ResNet-50]{\label{fig:compression:resnet50}\includegraphics[scale = 0.185]{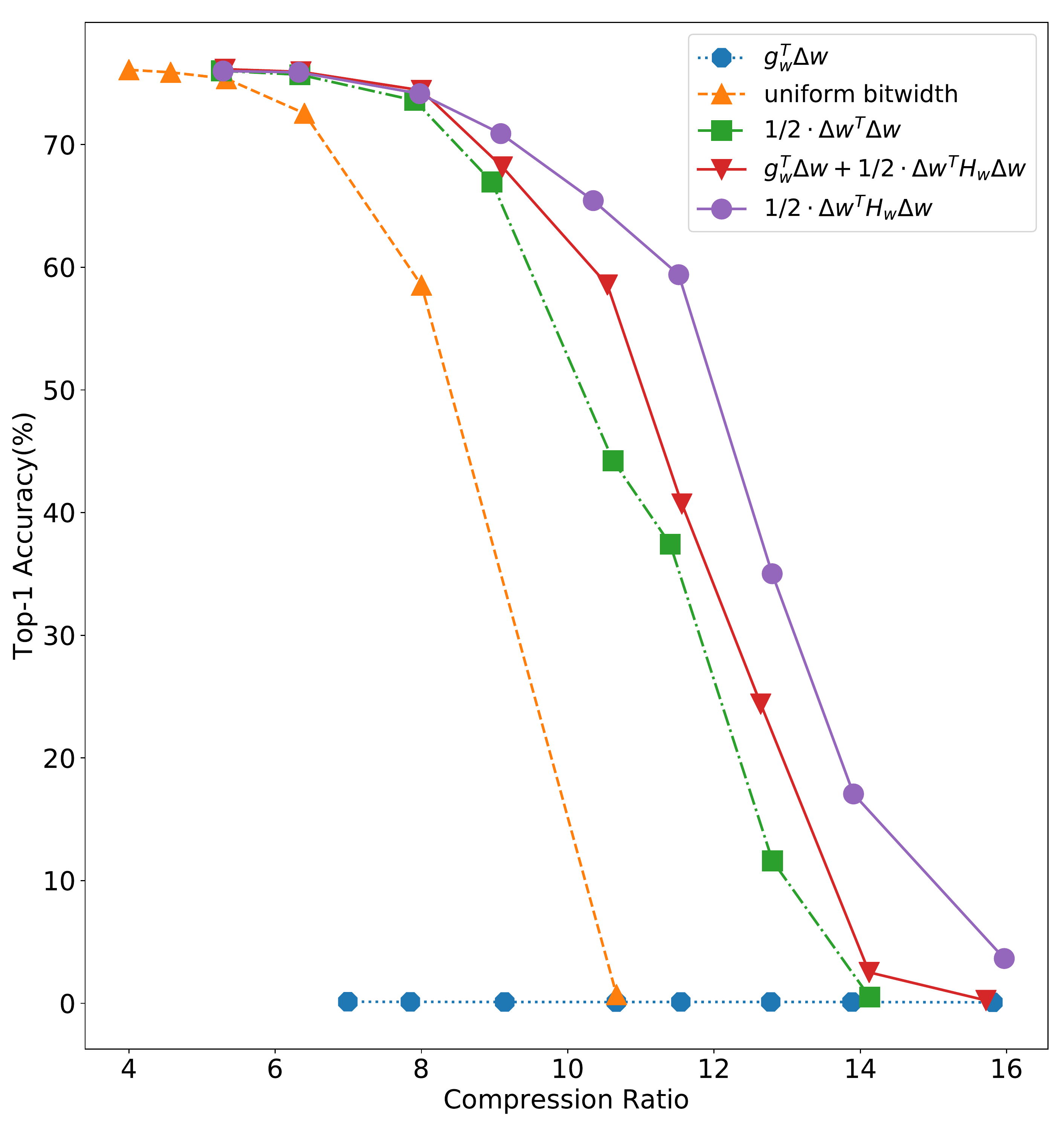}}
\subfigure[Mobilenet-V2]{\label{fig:compression:mobilenetv2}\includegraphics[scale = 0.185]{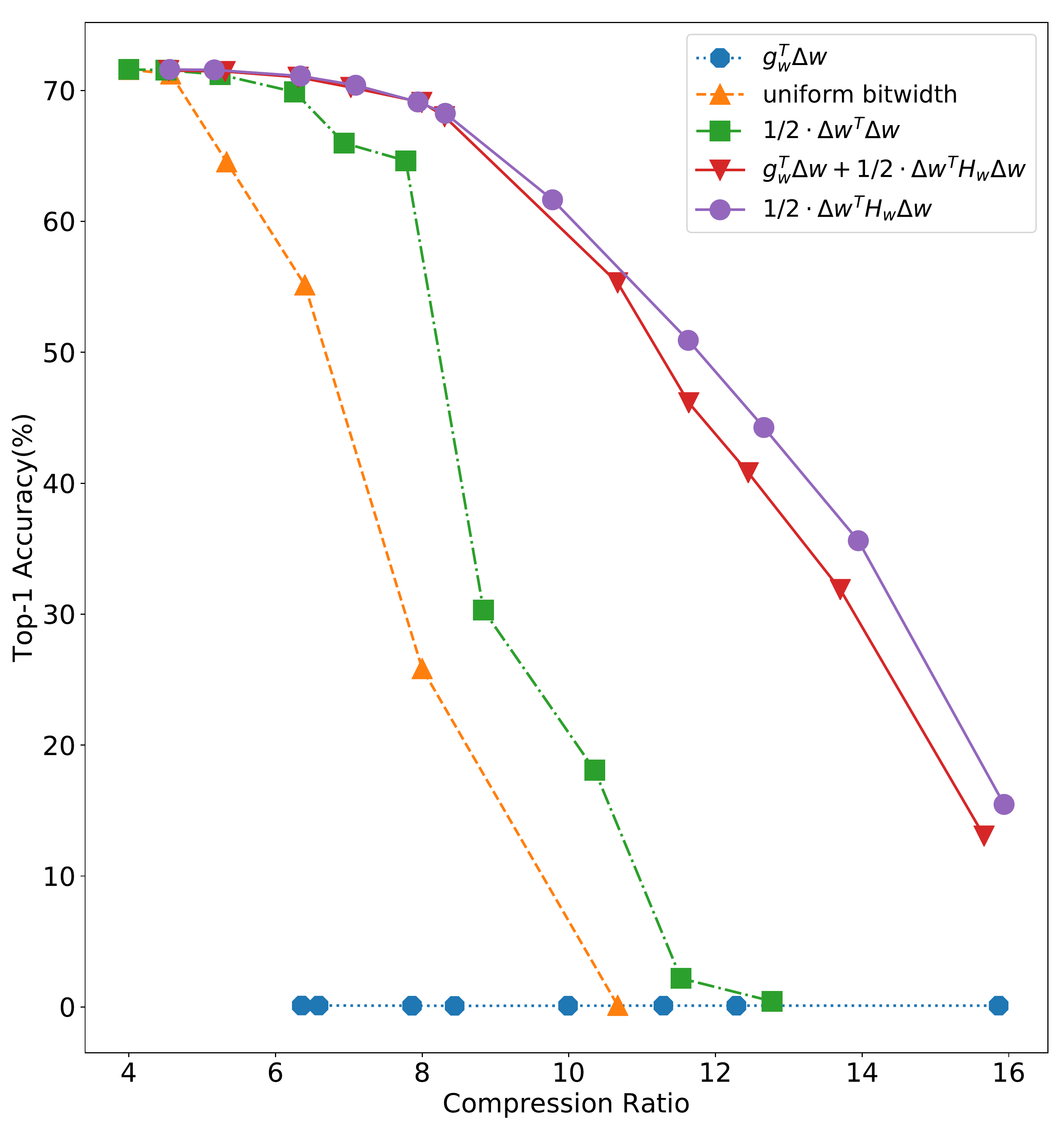}}
\caption{Comparision of different loss perturbation approximations for bit-width assignment of various kinds of architectures.}
\label{fig:compression}
\end{figure*}

\textbf{Approximated Hessian Matrix: } As stated above, we adopt the approximated second-order  term $\frac{1}{2}\Delta w^{T} \tilde{H}_{w} \Delta w$ as the proxy loss perturbation. 
To verify its advantages, we employ several other candidates for mixed-precision quantization and summarize the results in Figure \ref{fig:compression}. First, as the pre-trained model converges to a local minimum with nearly zero gradient vector, there is a devastating accuracy drop if only the first-order term is adopted. Second, compared with uniform bit-width and other Hessian-free (e.g. $\frac{1}{2}\Delta w^{T} \Delta w$) candidates, the significant performance improvement makes it worthwhile to pay for the extra computational cost of second-order information, especially for deep (e.g. ResNet-50) and lightweight (e.g. MobileNet-V2) networks. 
Finally, although it's theoretically better to combine the first and second-order terms, our experimental results contradict this intuition. 
We believe that it's because the extremely weak gradient information, which should be zero theoretically, acts more as noise for our loss perturbation approximation.

\begin{figure}[htbp]
\centering
\subfigure[ResNet-18]{\label{fig:bitwidth:resnet18}\includegraphics[scale = 0.16]{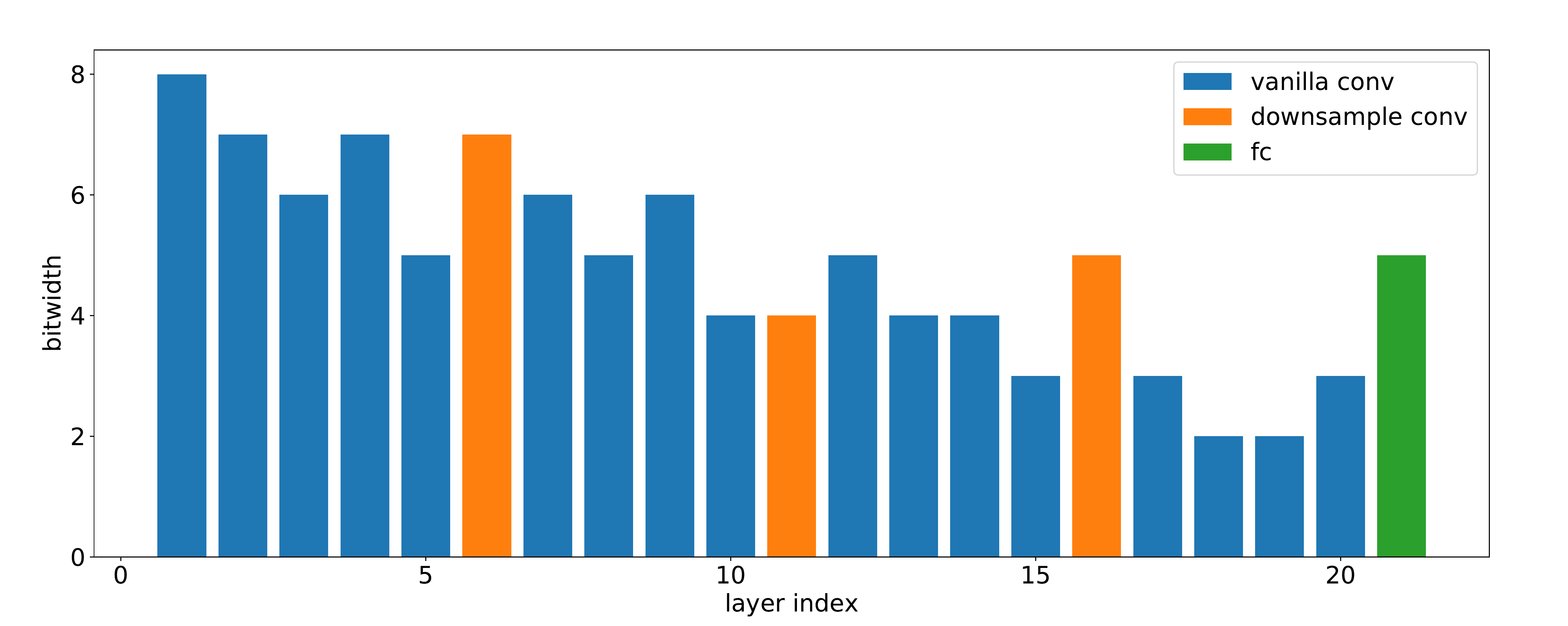}}
\subfigure[ResNet-50]{\label{fig:bitwidth:resnet50}\includegraphics[scale = 0.162]{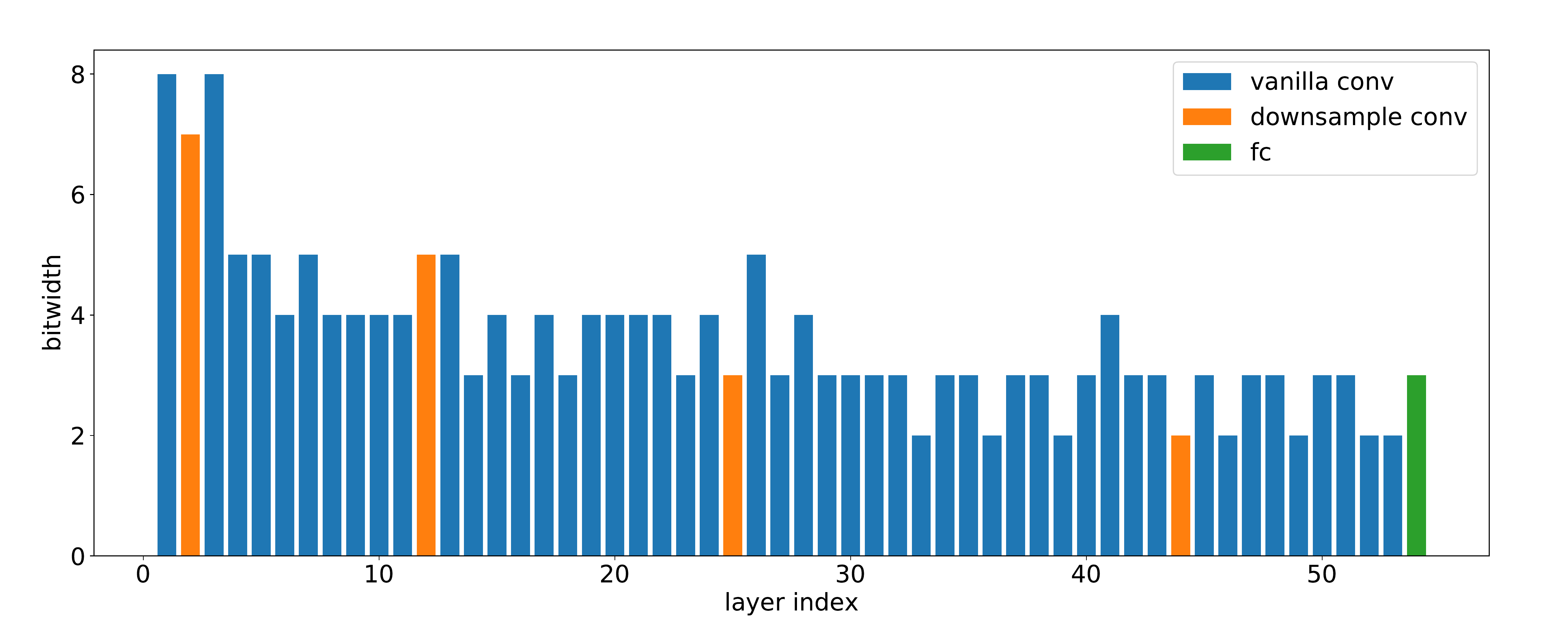}}
\subfigure[Mobilenet-V2]{\label{fig:bitwidth:mobilenetv2}\includegraphics[scale = 0.17]{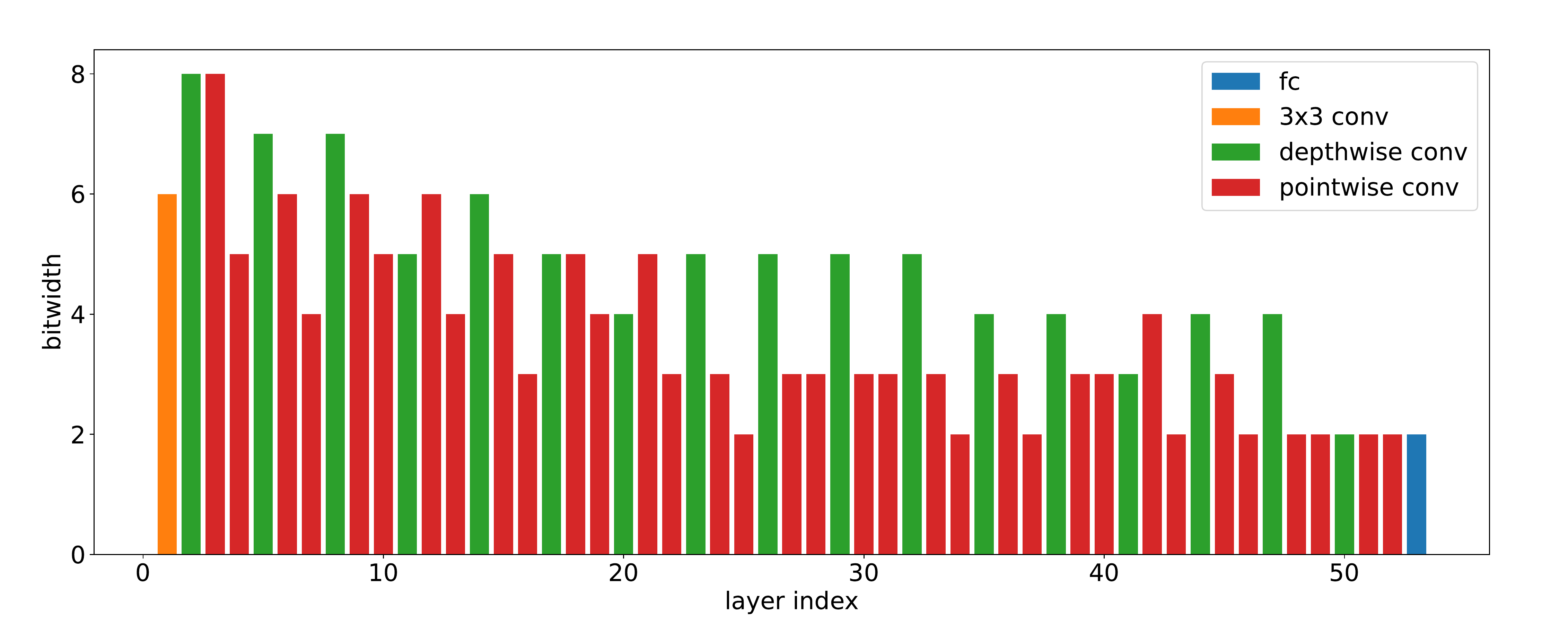}}
\caption{Bit-width assignment for various kinds of architectures.}
\label{fig:bitwidth}
\end{figure}

\textbf{MCKP Reformulation: } 
As MCKP is NP-hard, we propose a greedy search algorithm to solve it efficiently. To justify the effectiveness of the original criterion, we compare it with the other two ones, namely the reversed criterion and the random criterion.  
Specifically, the reversed criterion means that we choose the layer to increase bit-width with the lowest (instead of the highest) priority based on the original criterion, and the random criterion means that we choose the layer randomly. The results are summarized in Figure \ref{fig:greedy_ablation}.  
As we can see, the original criterion outperforms the other two ones consistently for various kinds of architectures.

\subsection{Comparision with SOTAs}
Furthermore, we compare the accuracy results after fine-tuning with several quantization methods proposed recently, which including uniform and mixed-precision quantization. The summarized results are reported in Table \ref{tab:summary_results}(\emph{see supplementary material for the experimental setup}).

For ResNet-18, compared with LQ-Nets \cite{lq-net} which introduces learnable scale factor for each bit, our method attains a smaller accuracy drop ($-0.26\%$ vs. $-0.30\%$) with larger compression ratio ($10.66\times$ vs. $6.10\times$). What's more, under the setting that the compression ratio of both weights and activations $\ge 8.00\times$, we can achieve almost lossless accuracy ($0.10\%$ drop), which improves significantly against other uniform quantization methods. 

For ResNet-50, except for uniform quantization, we also compare with other mixed-precision methods that including AutoQ \cite{autoq}, HAWQ \cite{hawq}, HAWQ-V2 \cite{hawq-v2}, and HAQ \cite{haq}. Compared with HAWQ and HAWQ-V2 that also utilize the second-order information of the model, we achieve significantly less accuracy drop ($-0.85\%$ vs. $-1.91\%$) with a similar compression ratio. Compared with HAQ that leverages reinforcement learning to search for optimal bit-width assignment, our method reaches the same accuracy drop ($0.85\%$ drop) with a much larger compression ratio for both weights and activations and less computation cost.

\begin{table*}[htbp]
\centering
\caption{Summary of quantization results on ImageNet dataset. We compare with various kinds of uniform quantization methods such as DC \cite{deepcomp}, ABC-Net \cite{abc-net}, LQ-Nets \cite{lq-net}, DoReFa-Net \cite{dorefa-net} and PACT \cite{pact}, and also recent mixed-precision methods such as AutoQ \cite{autoq}, HAWQ \cite{hawq}, HAWQ-V2 \cite{hawq-v2} and HAQ \cite{haq}. The `MP' refers to mixed-precision quantization, where we report the lowest bits used for weights and activations. The `w-ratio' and `a-ratio' stand for weight and activation compression ratio, respectively.}
\begin{threeparttable}[b]
\small
\begin{tabular}{|c|cccccccc|}
\hline
Network & Method & Top-1/Full & w-bits & a-bits & w-ratio & a-ratio & Top-1/Quant & Top-1/Drop \\ \hline \hline
\multirow{12}{*}{ResNet-18} & LQ-Nets\tnote{\dag} \ \cite{lq-net} & $70.30$ & $3$ & $32$ & $7.45\times$ & $1.00\times$ & $69.30$ & $-1.00$ \\
                           & LQ-Nets\tnote{\dag} \ \cite{lq-net} & $70.30$ & $4$ & $32$ & $6.10\times$ & $1.00\times$ & $70.00$ & $-0.30$ \\
                           & \textbf{Ours}    & $\bm{69.76}$ & $\bm{2_{MP}}$ & $\bm{32}$ & $\bm{10.66\times}$ & $\bm{1.00\times}$ & $\bm{69.50}$ & $\bm{-0.26}$ \\
                           & \textbf{Ours} 	 & $\bm{69.76}$ & $\bm{2_{MP}}$ & $\bm{8}$ & $\bm{10.66\times}$ & $\bm{4.00\times}$ & $\bm{69.39}$ & $\bm{-0.37}$ \\ \cdashline{2-9}
                           & ABC-Net \cite{abc-net} & $69.30$ & $5$ & $5$ & $6.40\times$ & $6.40\times$ & $65.00$ & $-4.30$ \\
                           & LQ-Nets\tnote{\dag} \ \cite{lq-net} & $70.30$ & $4$ & $4$ & $6.10\times$ & $7.98\times$ & $69.30$ & $-1.00$ \\
                           & DoReFa\tnote{\dag} \ \cite{dorefa-net} & $70.40$ & $5$ & $5$ & $5.16\times$ & $6.39\times$ & $68.40$ & $-2.00$ \\
                           & PACT\tnote{\dag} \ \cite{pact} & $70.40$ & $4$ & $4$ & $6.10\times$ & $7.98\times$ & $69.20$ & $-1.20$ \\
                           & \textbf{Ours} & $\bm{69.76}$ & $\bm{3_{MP}}$ & $\bm{4_{MP}}$ & $\bm{8.32\times}$ & $\bm{8.00\times}$ & $\bm{69.66}$ & $\bm{-0.10}$ \\ \hline \hline
\multirow{12}{*}{ResNet-50} & ABC-Net \cite{abc-net} & $76.10$ & $5$ & $5$ & $6.40\times$ & $6.40\times$ & $70.10$ & $-6.00$ \\
						   & LQ-Nets\tnote{\dag} \ \cite{lq-net} & $76.40$ & $3$ & $3$ & $5.99\times$ & $10.64\times$ & $74.20$ & $-2.20$ \\
                           & LQ-Nets\tnote{\dag} \ \cite{lq-net} & $76.40$ & $4$ & $4$ & $5.11\times$ & $7.99\times$ & $75.10$ & $-1.30$ \\
                           & DoReFa\tnote{\dag} \ \cite{dorefa-net} & $76.90$ & $4$ & $4$ & $5.11\times$ & $7.99\times$ & $71.40$ & $-5.50$ \\
                           & PACT\tnote{\dag} \ \cite{pact} & $76.90$ & $32$ & $4$ & $1.00\times$ & $7.99\times$ & $75.90$ & $-1.00$ \\
                           & PACT\tnote{\dag} \ \cite{pact} & $76.90$ & $2$ & $4$ & $7.24\times$ & $7.99\times$ & $74.50$ & $-2.40$ \\
                           & AutoQ\cite{autoq} & $74.80$ & $MP$ & $MP$ & $10.26\times$ & $7.96\times$ & $72.51$ & $-2.29$ \\
                           & HAWQ\cite{hawq} & $77.39$ & $2_{MP}$ & $4_{MP}$ & $12.28\times$ & $8.00\times$ & $75.48$ & $-1.91$ \\
                           & HAWQ-V2\cite{hawq-v2} & $77.39$ & $2_{MP}$ & $4_{MP}$ & $12.24\times$ & $8.00\times$ & $75.76$ & $-1.63$ \\
                           & HAQ\cite{haq} & $76.15$ & $MP$ & $32$ & $10.57\times$ & $1.00\times$ & $75.30$ & $-0.85$ \\ 
                           & \textbf{Ours} & $\bm{76.13}$ & $\bm{2_{MP}}$ & $\bm{4_{MP}}$ & $\bm{12.24\times}$ & $\bm{8.00\times}$ & $\bm{75.28}$ & $\bm{-0.85}$ \\ \hline \hline
\multirow{9}{*}{MobileNet-V2} & DC\cite{deepcomp} & $71.87$ & $MP$ & $32$ & $13.93\times$ & $1.00\times$ & $58.07$ & $-13.80$ \\
                              & HAQ\cite{haq} & $71.87$ & $MP$ & $32$ & $14.07\times$ & $1.00\times$ & $66.75$ & $-5.12$ \\
                              & \textbf{Ours} & $\bm{71.88}$ & $\bm{2_{MP}}$ & $\bm{8}$ & $\bm{13.99\times}$ & $\bm{4.00\times}$ & $\bm{68.52}$ & $\bm{-3.36}$ \\ \cdashline{2-9}
                              & DC\cite{deepcomp} & $71.87$ & $MP$ & $32$ & $9.69\times$ & $1.00\times$ & $68.00$ & $-3.87$ \\ 
                              & HAQ\cite{haq} & $71.87$ & $MP$ & $32$ & $9.69\times$ & $1.00\times$ & $70.90$ & $-0.97$ \\
                              & \textbf{Ours} & $\bm{71.88}$ & $\bm{2_{MP}}$ & $\bm{8}$ & $\bm{9.79\times}$ & $\bm{4.00\times}$ & $\bm{71.20}$ & $\bm{-0.68}$ \\ \cdashline{2-9}
                              & DC\cite{deepcomp} & $71.87$ & $MP$ & $32$ & $7.47\times$ & $1.00\times$ & $71.24$ & $-0.63$ \\ 
                              & HAQ\cite{haq} & $71.87$ & $MP$ & $32$ & $7.47\times$ & $1.00\times$ & $71.47$ & $-0.40$ \\
                              & \textbf{Ours} & $\bm{71.88}$ & $\bm{3_{MP}}$ & $\bm{8}$ & $\bm{7.49\times}$ & $\bm{4.00\times}$ & $\bm{71.83}$ & $\bm{-0.05}$ \\ \hline
\end{tabular}
\label{tab:summary_results}
\begin{tablenotes}
     \item[\dag] do not quantize the first and last layer
   \end{tablenotes}
  \end{threeparttable}
\end{table*}

At last, a much more efficient and lightweight architecture, MobileNet-V2, is utilized for further evaluation. Here we mainly compare with DC \cite{hansong} and HAQ \cite{haq}, which are uniform and mixed-precision quantization methods respectively. It should be noted that these two methods employ $k$-means algorithm to quantize the weights, and we instead adopt fixed-point quantization that sacrifices model accuracy for inference efficiency. 
Even under the situation of an unfair comparison, we still achieve significant performance improvement with a similar weight compression ratio and higher activation compression ratio in three different compression regimes, which justifies our method further.

\subsection{Bit-width assignment}
Finally, as shown in Figure \ref{fig:bitwidth}, we visualize the bit-width assignment for these three networks to understand what our method learns. 
First, on ResNet-18 and ResNet-50, as the first convolution layer processes the input image directly and is much lighter than other layers, it receives a higher bit-width.
Then, on ResNet-18, we notice that the output FC layer and downsample convolution layers also obtain higher bit-width, which is consistent with our prior knowledge that these components are critical for model performance. However, it should be noted that this conclusion does not hold strictly for ResNet-50, which is worthy of our further exploration. 
Besides, on MobileNet-V2, our method recognizes that depthwise convolution layers are more sensitive to quantization and allocates them higher bit-width, which is consistent with the conclusion of previous work\cite{haq}.

\section{Conclusion}
\label{sec:conclusion}
In this paper, we present a novel and principled framework to solve the mixed-precision quantization problem.
We first formulate the mixed-precision quantization as a discrete constrained optimization problem to provide a principled and holistic view. 
To solve the optimization problem, we propose an efficient approach to compute the Hessian matrix. Then we reformulate it as Multiple-Choice Knapsack Problem (MCKP) and propose a greedy search algorithm to solve it efficiently. 
Extensive experiments are conducted to demonstrate the efficiency and effectiveness of the proposed method over other uniform and mixed-precision quantization approaches.

  

\section*{Acknowledgement}
This work was supported in part by National Natural Science Foundation of China (No.61972396, No.61906193), National Key Research and Development Program of China (No. 2020AAA0103402), the Strategic Priority Research Program of Chinese Academy of Sciences (No. XDA27040300), the NSFC-General Technology Collaborative Fund for Basic Research (Grant No.U1936204).

{\small
\bibliographystyle{ieee_fullname}
\bibliography{egpaper_for_review}
}

\end{document}